\documentclass[preprint,review,3p,times,sort&compress,10pt]{elsarticle}

\usepackage{amssymb}
\usepackage{amsmath}

\usepackage{multirow}
\usepackage{orcidlink}
\usepackage{booktabs}
\usepackage{setspace}

\usepackage{colortbl}
\definecolor{firstcolor}{HTML}{FFC0CB}
\definecolor{secondcolor}{HTML}{FFF8C5}
\definecolor{thirdcolor}{HTML}{E2EEBC}

\newcommand{\fst}[1]{\cellcolor{firstcolor}\bfseries #1}
\newcommand{\snd}[1]{\cellcolor{secondcolor}#1}
\newcommand{\trd}[1]{\cellcolor{thirdcolor}#1}

\def\tsc#1{\csdef{#1}{\textsc{\lowercase{#1}}\xspace}}
\tsc{WGM}
\tsc{QE}
\tsc{EP}
\tsc{PMS}
\tsc{BEC}
\tsc{DE}

\journal{Elsevier}

\begin{document}

\begin{frontmatter}



\title{Hadamard Attention Recurrent Transformer: A Strong Baseline for Stereo Matching Transformer} 

\author[gzu]{Ziyang Chen\orcidlink{0000-0002-9361-0240}}

\author[gzcc]{Wenting Li\orcidlink{0009-0009-8081-842X}}

\author[gzu]{Yongjun Zhang\orcidlink{0000-0002-7534-1219}\corref{mycorrespondingauthor}}
\cortext[mycorrespondingauthor]{Corresponding author.}

\author[gzu]{Yabo Wu\orcidlink{0009-0007-2504-0219}}

\author[npu]{Bingshu Wang\orcidlink{0000-0002-2603-8328}}

\author[gzu,pku]{Yong Zhao\orcidlink{0000-0002-7999-1083}}

\author[scut]{C. L. Philip Chen\orcidlink{0000-0001-5451-7230}}

\affiliation[gzu]{organization={College of Computer Science, the State Key Laboratory of Public Big Data},
             addressline={Guizhou University},
             city={Guiyang},
             postcode={550025},
             country={China}}
\affiliation[gzcc]{organization={School of Information Engineering},
            addressline={Guizhou University of Commerce},
            city={Guiyang},
            postcode={550021},
            country={China}}
        
\affiliation[npu]{organization={School of Software},
			addressline={Northwestern Polytechnical University},
			city={Xi'an},
			postcode={710129},
			country={China}}
		
\affiliation[pku]{organization={Key Laboratory of Integrated Microsystems, Shenzhen Graduate School},
	addressline={Peking University},
	city={Shenzhen},
	postcode={518055},
	country={China}}
        
\affiliation[scut]{organization={School of Computer Science and Engineering},
	addressline={South China University of Technology},
	city={Guangzhou},
	postcode={510641},
	country={China}}



\begin{abstract}
Constrained by the low-rank bottleneck inherent in attention mechanisms, current stereo matching transformers suffer from limited nonlinear expressivity, which renders their feature representations sensitive to challenging conditions such as reflections. To overcome this difficulty, we present the \textbf{H}adamard \textbf{A}ttention \textbf{R}ecurrent Stereo \textbf{T}ransformer (HART). HART includes a novel attention mechanism that incorporates the following components: 1) The Dense Attention Kernel (DAK) maps the attention weight distribution into a high-dimensional space over (0, +$\infty$). By removing the upper bound constraint on attention weights, DAK enables more flexible modeling of complex feature interactions. This reduces feature collinearity. 2) The Multi Kernel \& Order Interaction (MKOI) module extends the attention mechanism by unifying semantic and spatial knowledge learning. This integration improves the ability of HART to learn features in binocular images. Experimental results demonstrate the effectiveness of our HART. In reflective area, HART ranked \textbf{1st} on the KITTI 2012 benchmark among all published methods at the time of submission. Code is available at \href{https://github.com/ZYangChen/HART}{here}.
\end{abstract}


\begin{highlights}
\item Hadamard Attention Recurrent Stereo Transformer (HART) is an efficient stereo transformer based on the hadamard product. The attention mechanism of HART is designed to be scalable, positioning HART as a potential new baseline for stereo transformer models.
\item To enhance matching performance under challenging conditions, we propose the Dense Attention Kernel (DAK) to address the low-rank bottleneck in transformers. DAK improves the nonlinear expressiveness of the model, enabling a better understanding of disparity relationships in complex scenarios.
\item To enhance spatial and channel interactions, we introduce the Multi-Kernel \& Order Interaction (MKOI) module, designed to capture both global and local features.
\item HART ranked \textbf{1st} on the KITTI 2012 reflective leaderboard and and secured a position in the \textbf{Top ten} on the Middlebury benchmark among all published methods at the time of submission. 
\end{highlights}


\end{frontmatter}



\section{Introduction}
\doublespacing
Stereo matching represents a fundamental challenge in the field of computer vision, with a range of practical applications including autonomous driving, remote sensing, robotics and augmented reality \cite{chen2024surface,gan2025prior,llk}. The goal of stereo matching is to compute the pixel-by-pixel correspondences between two images, and to generate a disparity map that can be translated into depth using the settings of the stereo camera system. In practical, it needs to achieve excellent matching capabilities in a variety of scenarios. 

\begin{figure}[]
	\centering
	\includegraphics[width=0.5\linewidth]{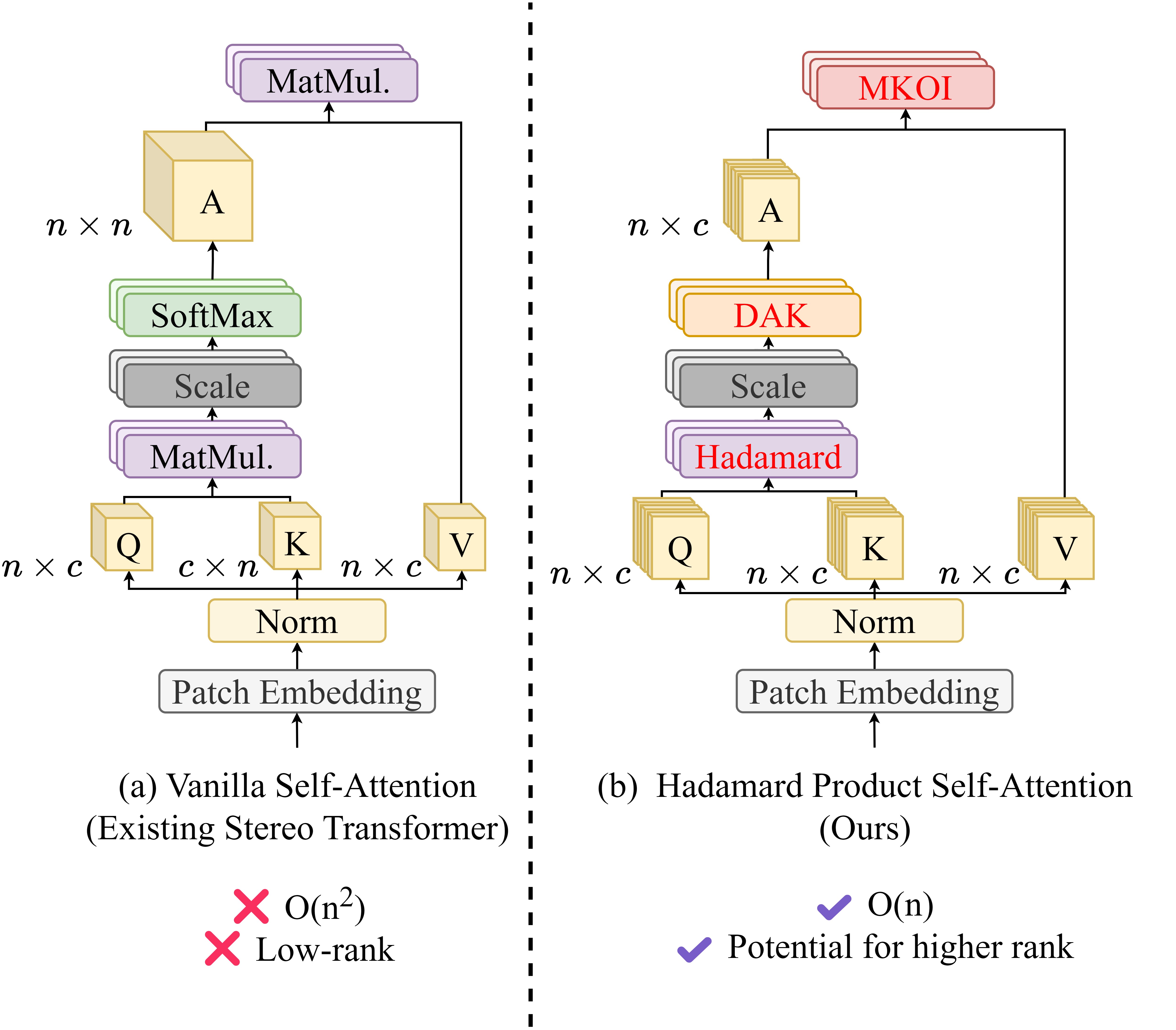}
	\caption{\textbf{Motivation.} The ability to effectively and efficiently express in ill-posed regions, i.e. reflective area, is a crucial aspect of the stereo matching application. The low-rank bottleneck and the quadratic time complexity present significant challenges for stereo transformers in meeting the aforementioned needs. Our objective is to address these issues through the implementation of HART.
	} 
	\label{motivation}
\end{figure}

Since the development of deep learning, numerous stereo matching Convolutional Neural Networks (CNNs) have been presented. Recently, recurrent-based methods have shown promising performance on stereo matching benchmarks. RAFT-Stereo \cite{raftstereo} uses a Recurrent All-pairs Field Transform (RAFT) paradigm based on multi-level convolutional Gated Recurrent Units (convGRU). IGEV-Stereo \cite{igev} proposes the Combined Geometry Encoding Volume (CGEV) instead of the All-pairs Correlation (APC) Volume \cite{raftstereo} in RAFT-Stereo. MoCha-Stereo \cite{mocha} mining the recurrent feature in the feature channels and achieves state-of-the-art (SOTA) performance. 
While the limited receptive field restricts the learning of long-range information. 

The development of Vision Transformer (ViT) has addressed the issue of acquiring long-range dependencies. STTR \cite{sttr} uses a transformer architecture, replacing cost volume construction with dense pixel matching from a sequence-to-sequence correspondence perspective. ELFNet \cite{elfnet} combines stereo transformers with a multi-scale cost volume. 
However, existing stereo transformers still have certain limitations. 

As show in Fig. \ref{motivation}, stereo transformers are still constrained by the quadratic-level complexity of attention, limiting the speed of inference. They are challenging to meet the efficient computational requirements of scenarios such as autonomous driving. Moreover, vanilla self-attention (SA) also falls into a low-rank bottleneck \cite{lowrank}. 
Specifically, each query matrix $Q$ and key matrix $K$ for every attention head jointly comprise only $\frac{2nc}{d}$ parameters, where $n = h \times w$ represents the spatial dimension, $h$ and $w$ denote the height and width of feature maps separately, $c$ represents the channel dimension, $d$ means the number of attention heads. Meanwhile, the attention matrix $A = Q \times K$, where the number of parameters is $n^2$, satisfies  $\frac{2nc}{d} << n^2$. 
This implies that stereo transformers have limited nonlinear expressive power, making them prone to being affected by ill-posed regions, such as reflections and weak textures.

To alleviate the above problems, \textbf{H}adamard \textbf{A}ttention \textbf{R}ecurrent Stereo \textbf{T}ransformer (HART), a new baseline for stereo transformer is proposed. Our primary contributions can be summarized as follows: 

1) 
To achieve an efficient architecture, we propose a novel stereo transformer pipeline based on the Hadamard product. This approach ensures that the computation is performed with linear time complexity.

2) 
To enhance matching performance under challenging conditions, we propose the Dense Attention Kernel (DAK) to address the low-rank bottleneck in transformers. DAK improves the nonlinear expressiveness of the model, enabling a better understanding of disparity relationships in complex scenarios.

3) 
To enhance spatial and channel interactions, we introduce the Multi-Kernel \& Order Interaction (MKOI) module, designed to effectively capture both global and local features.

4) 
HART ranked \textbf{1st} on the KITTI 2012 reflective leaderboard and and secured a position in the \textbf{Top ten} on the Middlebury benchmark among all published methods at the time of submission. Moreover, the attention mechanism of HART is designed to be scalable, positioning HART as a potential new baseline for stereo transformer models.

\section{Related Work}
\subsection{Recurrent-based Stereo CNNs}
Learning-based methods \cite{gwcnet,mocha,fdn} have grown rapidly in recent years. Among them, recurrent-based algorithms stand out for their performance in real-world scenarios. As a representative of recurrent-based techniques, RAFT-Stereo \cite{raftstereo} validates the effectiveness of the RAFT iteration and the APC pyramid structure. Subsequent improvements mainly focus on the computation of matching costs in the stereo matching field. CREStereo \cite{cre} designs Adaptive Group Correlation Layer (AGCL) to compute the correlation of each feature map at different cascade levels separately, refining the differences independently through multiple iterations. 
IGEV-Stereo \cite{igev} computes an additional set of Geometry Encoding Volumes (GEV) for correlation caculation. MoCha-Stereo \cite{mocha} restores geometric details lost in the feature channels and calculates a correlation based on recurrent patterns within the channels. This process results in a new cost volume termed Motif Channel Correlation Volume (MCCV). The utilization of MCCV has helped recurrent-based stereo CNNs achieve SOTA performance on many datasets. 

Despite the leading performance of such methods, recurrent-based schemes typically use convolutional operations with relatively small receptive fields. They are not conducive to learning global information \cite{fsa}. This makes recurrent-based CNNs challenging in areas where the semantic interpretation cannot be determined from local geometric information alone.

\subsection{Stereo Transformers}
Benefiting from the global learning capability of ViT, stereo transformers \cite{sttr,elfnet,croco} deliver excellent performance. STTR \cite{sttr} constructs a transformer module with alternating self-attention and cross-attention to compute feature descriptors. ELFNet \cite{elfnet} utilizes self-attention transformer to learn global features as part of estimating evidence-based disparities. 
CroCo-Stereo \cite{croco} stacks SA and MLP-designed encoder and decoder. Since both approaches incorporate vanilla self-attention (SA) within their transformer blocks, existing stereo transformers struggle to match ill-posed regions such as reflective surfaces and thin structures.

Though the design of the transformer allows for long-range matching \cite{wyb1}, it is a computationally expensive model. SA has quadratic time complexity. 
As a result, existing stereo transformers generally require longer inference times. Currently, there are efforts to reduce the time complexity of stereo transformers. DLNR \cite{dlnr} and RetinaStereo \cite{retina} borrow ideas from the attention in Restormer \cite{restormer,hjl} and the recurrent update in RAFT-Stereo \cite{raftstereo}. It deviates from the vanilla spatial feature matmul product and instead designs a channel feature matmul product. However, like other recurrent-based methods, DLNR and RetinaStereo lack the spatial interactions of vanilla SA. In addition, although the number of channels $c$, is much smaller than $h \times w$, $c^2$ is still much larger than $h \times w$ in DLNR and RetinaStereo. This means that the SA computation of DLNR and RetinaStereo \cite{dlnr,retina} still have a time complexity of $O(c^2)$.

In summary, stereo transformers still face two main challenges: 1) Inaccurate detail due to low-rank bottleneck; 2) High computational cost of SA. 
We hypothesize that the low-rank bottleneck inherent in vanilla SA restricts the expressive capacity of stereo transformers. The limited nonlinear expressiveness of SA makes it difficult for the model to capture homonymy point correspondences in ill-posed regions. Additionally, to improve computational efficiency, we aim to reduce the complexity of the attention mechanism from $O(n^2)$ to $O(n)$.

\subsection{Efficient Transformers for Vision Tasks}
In other tasks where transformers \cite{transformer} are applied, numerous studies \cite{linformer,eff-att} have attempted to achieve linear attention. The implementation of linear transformers typically decouples the SoftMax operation and altering the calculation order to first compute $K^T \times V$. Nevertheless, this form of attention matrix also suffers from the low-rank problem \cite{lowrank}.
Each query matrix $Q$ and key matrix $K$ for every attention head jointly comprise only $\frac{2nc}{d}$ parameters, where $n = h \times w$ represents the spatial dimension, $h$ and $w$ denote the height and width of feature maps separately, $c$ represents the channel dimension, $d$ means the number of attention heads. Meanwhile, the attention matrix $A = Q \times K$, where the number of parameters is $n^2$, satisfies  $\frac{2nc}{d} << n^2$. The low-rank bottleneck results in a decrease in expressiveness \cite{notpure}. 

\begin{figure*}[]
	\centering
	\includegraphics[width=\linewidth]{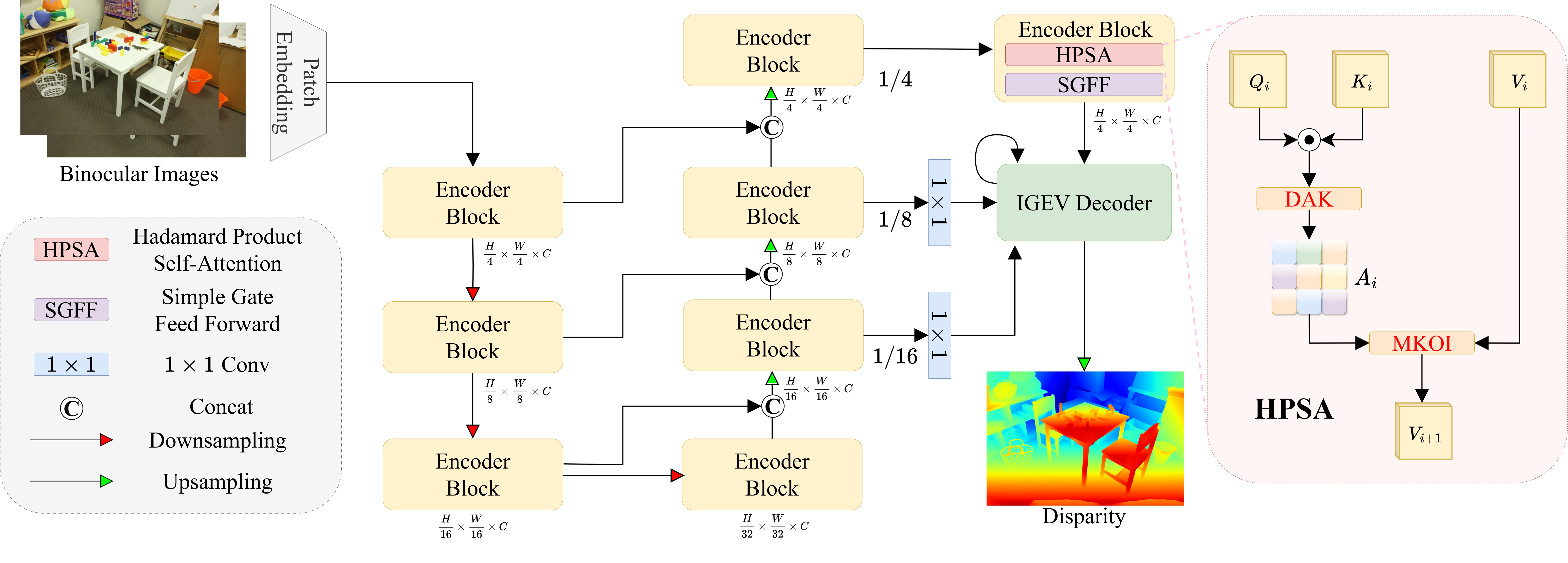}
	\caption{The overall architecture of our HART. We construct multi-scale transformer blocks to encode features. The attention and forward propagation of the features are implemented by HPSA and SGFF respectively. We design the IGEV Decoder with reference to IGEV-Stereo. Based on the iterative updating paradigm and correlation volume, adaptations are made to the IGEV Decoder to suit our Stereo Transformer. } 	
	\label{pipeline}
\end{figure*}

\section{Method}
\subsection{Overview}
We designed Hadamard Attention Recurrent Transformer (HART), a strong baseline for stereo transformers. We believe that improving the matching performance of stereo transformers hinges on overcoming the low-rank bottleneck \cite{lowrank} of SA, and incorporates a series of new components: 1) DAK broadens the range of activation values, thereby reducing feature similarity. This mechanism alleviates the expressiveness degradation caused by the low-rank bottleneck, which facilitates the HART in matching homonymous points within ill-posed regions. 2) MKOI is proposed to supplement the spatial and cross-channel interaction capabilities lacking in the Hadamard product. 

DAK and MKOI jointly constitute the novel attention mechanism we propose, referred to as Hadamard Product Self-Attention (HPSA). The HPSA process is described in Sections \ref{matrix_a} and \ref{mkoi}. To complete the HART workflow, we introduce techniques for iterative disparity updating and correlation computation as part of the decoder. The full architecture of HART is portrayed in Fig. \ref{pipeline}. The Simple Gate Feed Forward (SGFF) mechanism is detailed in Section \ref{sgff}. HPSA and SGFF form the encoder of HART, the decoder of HART is outlined in Section \ref{igevdecoder}. 
Inspired by STTR \cite{sttr} and IGEV-Stereo \cite{igev}, we supervised HART by the Equ. \ref{loss}, where $\gamma=0.9$, $N$ denotes the total number of iterations, $d_0$ represents the initial disparity computed from correlation volume in our IGEV Decoder, and the loss for the initial disparity is calculated using smooth L1. L1 loss is used for subsequent iteration processes.
\begin{equation}
	L=Smooth_{L1}(d_0-d_{gt})+\sum_{i=1}^{N}\gamma ^{N-i}||d_i-d_{gt}||_1
	\label{loss}
\end{equation}

\subsection{Caculation of Attention Matrix}
\label{matrix_a}

Although stereo transformers can achieve good results in long-range matching, this architecture is still not efficient enough. First, stereo transformers often produce incorrect matches for homonymous points in ill-posed regions. We attribute this issue to the low-rank bottleneck inherent in vanilla SA. Secondly, the quadratic complexity of the attention mechanism makes it difficult for transformers to achieve efficient performance. 
To achieve linear time complexity, HPSA is designed to be used here. We implemented a series of designs to prevent HPSA from falling into the low-rank bottleneck. 

\subsubsection{Hadamard Product between Query and Key}

Vanilla SA in the stereo transformer \cite{sttr,elfnet,croco} relies on matmul product to compute the attention matrix $A$. However, this computational approach has a time complexity of $O(n^2)$. $n = h \times w$ here, $h$ denotes the height, $w$ means width of the feature map. $Q,K,V,A$ stand for query, key, value and attention matrix respectively. In existing stereo transformers, it holds that $n >> c$, where $c$ means the number of feature channels. The above conclusion is a proof that the vanilla SA exhibits a quadratic level of complexity. Despite the introduction of channel-wise SA \cite{restormer,dlnr} for stereo transformers, the condition $c^2 >> n$ still prevails. To improve the computational efficiency of attention, we compute the attention matrix using the Hadamard product, as illustrated in Equ. \ref{hada}. 
\begin{gather}
	A = ||Q||_2 \odot ||K||_2 \label{hada}	\\
	O(A) = O(Q_{c \times n} \odot K_{c \times n}) = O(c \cdot n) \triangleq O(n)	\label{thada}
\end{gather}
where $\odot$ refers to the Hadamard product. The attention matrix $A$ is computed solely through the Hadamard product of $Q$ and $K$. According to Equ. \ref{thada}, our attention calculation realises linear time complexity. 

\subsubsection{Dense Attention Kernel (DAK)}
Attention matrix $A$ comprises only $n \cdot c$ parameters, and the $Q$ and $K$ matrices together consist of $2\cdot n\cdot c$ parameters. This implies that, in theory, each parameter can be adequately represented. To achieve this, we aim for all parameters to maintain a non-zero state. Following this rationale, we have designed the Dense Attention Kernel (DAK), as depicted in Equ. \ref{kernel}. 
\begin{gather}
	DAK(A)=\left\{
	\begin{aligned}
		A + 1, A & \geq & 0 \\
		e^{A}, A & \textless & 0
	\end{aligned}
	\label{kernel}
	\right.
\end{gather}
In comparison to the SoftMax utilized in vanilla SA and ELU used as activate function, our DAK always satisfied $DAK(A) \textgreater 0$. 
By extending the activation range of attention from (0, 1) or (-1,+$\infty$) to (0, +$\infty$), the Dense Attention Kernel (DAK) mitigates the issue of silent neurons and promotes a higher effective matrix rank through the conversion of zero elements into non-zero ones. This transformation reduces feature similarity and helps alleviate the low-rank bottleneck, ultimately enhancing the model's nonlinear representational capacity. 
The derivative properties of DAK determine its ability to further widen the attentional gap, thereby redirecting the attention to more important features. The broader range of attentional weighting contributes to a nuanced emphasis on key features. 
The $DAK(A)$ serves as a weight in the description of the characteristics for the Value $V$. We can ensure that each dimension is positively correlated with the weights here. Because the feature values of $A$ at a given pixel are described by the channel features of that pixel. 

In addition to amplifying the differences in attentional response scores, DAK also shows more robust expressiveness than ELU. 
This is attributed to the fact that DAK can be decoupled into a process involving Equ. \ref{attelu}.
\begin{gather}
	V_{i+1}=MKOI(DAK(A_i)\odot V_{i})
	\triangleq MKOI(V_{i}+ELU(A_i)\odot V_{i})
	\label{attelu}
\end{gather}
where $V_{i}$ and $V_{i+1}$ means Value before and after our attention processing, and $ELU$ means ELU function. (See text in next section for content of MKOI.) Based on this formula, we can derive the gradient computation formula for the learning process of attention from shallow attention $i$ to the deeper position $I$ \cite{resnet}. This process takes the form of Equ. \ref{lossgra}. 
\begin{gather}
	\frac{ \partial loss }{ \partial V_{i} }=
	\frac{ \partial loss }{ \partial V_{I} } \cdot
	\frac{ \partial V_{I} }{ \partial V_{i} }
	\triangleq \frac{ \partial loss }{ \partial V_{I} } \cdot MKOI(1+
	\frac{ \partial  }{ \partial V_{i} } \sum_{j=i}^{I-1} (ELU(A_j) \cdot V_j)
	)
	\label{lossgra}
\end{gather}
These ensure more diverse feature representation, theoretically allowing for more precise matching.

\begin{figure}[]
	\centering
	\includegraphics[width=\linewidth]{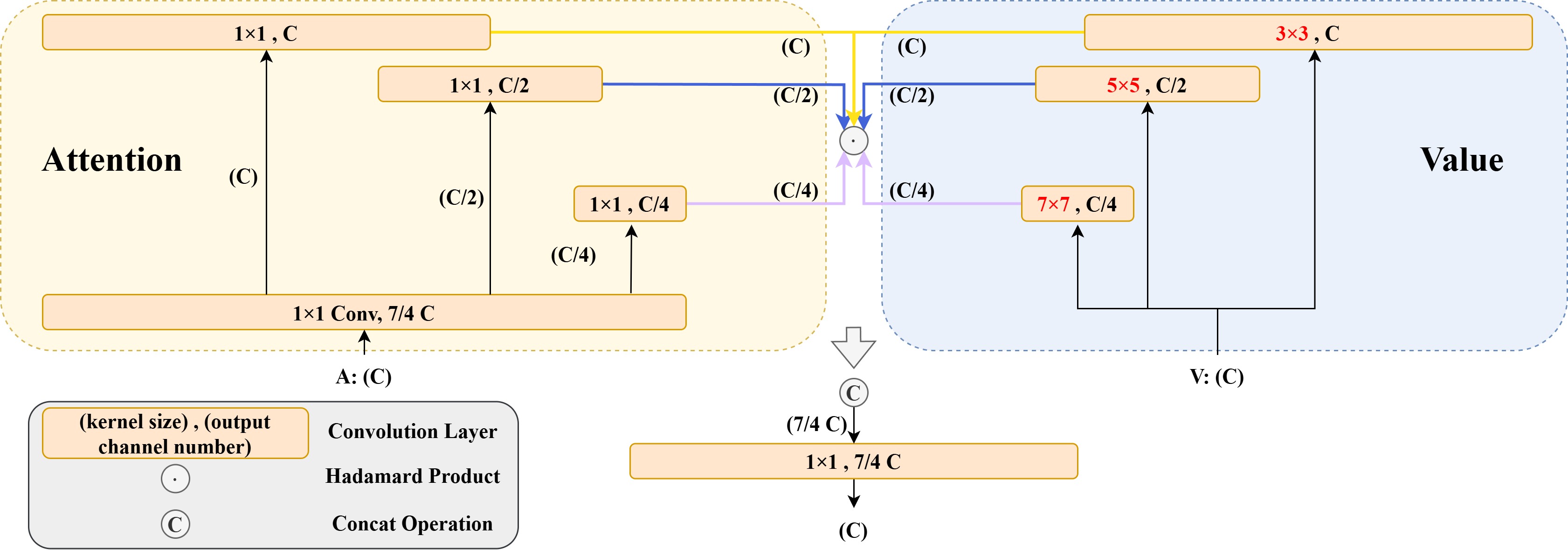}
	\caption{Overview of \textbf{M}ulti \textbf{K}ernel \& \textbf{O}rder \textbf{I}nteraction (MKOI). The left box shows the splitting and convolution of the Attention matrix $A$, and the right box shows the splitting and convolution of the Value matrix $V$. It is worth noting that the letters in the brackets represent the number of channels. Lines of the same color represent the corresponding Hadamard product. } 
	\label{MKOI}
\end{figure}
\subsection{Multi Kernel \& Order Interaction (MKOI)}
\label{mkoi}
Vanilla self-attention (SA) satisfied $SA(Q,K,V)=SoftMax(\frac{Q K^T}{\sqrt{d_k}})V$. 
This process involves spatial and channel interactions. Vanilla Hadamard product cannot achieve this. To improve the spatial and channel interaction of HPSA, we propose the Multi Kernel \& Order Interaction (MKOI) module as a complement to HPSA. 

As shown in Fig. \ref{MKOI}, MKOI is a Hadamard product paradigm concerning the Attention matrix $A$ and the Value matrix $V$. To supplement the channel interaction capability, MKOI decouples features into four sets with channel numbers $[c, c/2, c/4, c/4]$, where $c$ represents the original number of channels. The rough semantic distributions of the image are already implied by lower order features \cite{CNNvisualizing}. Therefore, we use large kernel convolutions to extract global semantic distributions. Compared to applying large kernel convolutions to all channels \cite{large}, this extraction mode is obviously cheaper. Higher-order features can focus on local features, so we use 1$\times$1 and 3$\times$3 convolutions to extract details. Through this paradigm, MKOI can capture both long-range and short-range information.

As a similar work, $g^nConv$ \cite{hornet} is also a channel-wise operation. MKOI differs from $g^nConv$ mainly in two aspects. 1) $g^nConv$ only performs convolution within a specified kernel size, making it difficult to simultaneously focus on both global and local information. 2) The channel interaction of $g^nConv$ relies solely on the dot product operations between split features, resulting in a bottleneck in channel information interaction. MKOI achieves channel interaction between Attention features and Value features, enhancing attention to the channels.
\begin{gather}
	HPSA(Q,K,V)= w^{\frac{7}{4}c \rightarrow c}_{1 \times 1}(
	[\underline{
		DAK(||Q^{\frac{c}{2^m}}||_2\odot ||K^{\frac{c}{2^m}}||_2) \odot w^{c \rightarrow \frac{c}{2^m}}_{s \times s}(V)
	}]_{m=\{0,1,2\}})
	\label{MKOIHPSA} \\
	V_{i+1} = HPSA(Q_i,K_i,V_i)
\end{gather}

Overall, the HPSA combined with MKOI can be expressed by Equ. \ref{MKOIHPSA} and Fig. \ref{motivation} (b). MKOI uses 1 $\times$ 1 convolutions to compress and interact channels, producing $V_{i+1}$ as input to the next transformer block.
$[\underline{\dots}]$ means concat along the feature channel dimension here, $s$ denotes the kernel size and $s=2m+3$.

\subsection{Simple Gate Feed Forward (SGFF)}
\label{sgff}
To complete the transformer architecture and further improve nonlinear expressiveness, we introduce a forward propagation architecture with reference to \cite{restormer}. This process can be written as Equ.\ref{SGFF1} and Equ. \ref{SGFF2}.
\begin{gather}
	Gate(V_{i+1}) = GELU(w^{c \rightarrow c}_{3 \times 3}(V_{i+1}^{in})) \odot w^{c \rightarrow c}_{3 \times 3}(V_{i+1}^{in})	\label{SGFF1}	\\
	SGFF(V_i) = w^{c \rightarrow c}_{1 \times 1}(Gate(w^{c \rightarrow c}_{1 \times 1}(V_{i+1})))
	\label{SGFF2}
\end{gather}
where $GELU$ means GELU operation. The overall computation process of our transformer block can be represented as Equ. \ref{block1} and Equ. \ref{block2}.. 
\begin{gather}
	Block_{att}(V^{init}) = V^{init} + HPSA(Q,K,V)	\label{block1} \\
	V_{i+1} = Block_{att}(V_i) + SGFF(LN(Block_{att}(V_i)))
	\label{block2}
\end{gather}
where $Block$ means the our transformer block in Fig. \ref{pipeline}, $LN$ denotes to Layer Normalization \cite{ln}. 
$Q$, $K$, and $V$ here are three variables derived from the initial feature $V^{init}$. A 1$\times$1 convolutional layer is employed to produce features $V^{3c}$ with a channel size of $3c$. These features are then evenly split into three parts along the channel dimension, representing $Q$, $K$, and $V$ respectively. 

\subsection{IGEV Decoder}
\label{igevdecoder}
The encoding structure of HART already demonstrates excellent capabilities, retaining long-term memory also aids in enhancing the disparity decoding abilities. We have therefore replaced the convGRU recurrent updating operator with LSTM-structured, inspired by \cite{fdn,dlnr,mocha,v2}. 
Outputs of this iterative update operation are added as increments to the previously calculated disparity values from the $i$-th iteration. 
For correlation volume, we follow IGEV-Stereo \cite{igev} to compute a group-wise correlation volume, and refine it with a regularization network $\mathbb{R}$. 
\begin{gather}
	\mathbb{C}_{GEV}(d,h,w,g) = \mathbb{R}(\frac{1}{N_c/N_g}\langle
	f^g_{l,4}(h,w),f^g_{r,4}(h,w+d)\rangle, F(l))
	\label{gwc}
\end{gather}
where $N_c$ is the number of channels, $N_g$ means the group number ($N_g=8$ here), $d$ denotes to the disparity level,  $\langle \cdots,\cdots \rangle$ is the inner product, $F$ is a frozen network pretrained by \cite{efficientnet}, $l$ is the left view image. The construction of 3D regularization network $\mathbb{R}$ also follows IGEV-Stereo \cite{igev}.

\begin{table}[]
	\centering	
	\caption{Default hyperparameters of HART.}
	\label{hyper}
	\begin{tabular}{ll}
		\hline
		use mixed precision & True \\
		batch size used during training & 8 \\
		crop size & 320 $\times$ 720 \\
		max learning rate & 0.0002 \\
		length of training schedule & 200000 \\
		recurrent-number during training & 22 \\
		Weight decay in optimizer & 0.00001 \\
		recurrent-number during evaluation & 32 \\
		number of levels in the correlation pyramid & 2 \\
		width of the correlation pyramid & 4 \\
		resolution of the disparity field (1/2\textasciicircum{}K) & 2 \\
		number of hidden recurrent levels & 3 \\
		max disp of correlation encoding volume & 192 \\
		color saturation & {[}0, 1.4{]} \\
		random seed & 666\\
		\hline
	\end{tabular}
\end{table}

\section{Experiment}
\subsection{Implementation Details}
The PyTorch framework is used to implement HART. The AdamW optimizer is used during training. Tab. \ref{hyper} collates the hyperparameters used by HART. The KITTI 2012 \cite{kitti2012}, KITTI 2015 \cite{kitti2015}, Scene Flow \cite{sceneflow}, and Middlebury \cite{middlebury} datasets were used to assess the performance of our model. 
HART is trained for $200,000$ steps on the Scene Flow \cite{sceneflow} dataset with a batch size of $8$. The input photographs are cropped at random to 320 $\times$ 736 pixels. We use data augmentation methods including spatial transformations and asymmetric chromatic augmentations. 
For training on the Middlebury dataset \cite{middlebury}, we begin by fine-tuning the Scene Flow \cite{sceneflow} pretrained model with a crop size of 384 $\times$ 512 for 200,000 steps, using the mixed Tartan Air \cite{tartanair}, CREStereo \cite{cre}, Scene Flow \cite{sceneflow}, Falling Things \cite{falling}, InStereo2k \cite{instereo2k}, CARLA \cite{carla}, and Middlebury \cite{middlebury} datasets. We then use a batch size of 8 for an additional 100,000 steps to refine the model on the mixed CREStereo dataset \cite{cre}, Falling Things \cite{falling}, InStereo2k \cite{instereo2k}, CARLA \cite{carla}, and Middlebury datasets \cite{middlebury}. 
For training on the KITTI 2012 \cite{kitti2012} and KITTI 2015 \cite{kitti2015} datasets, we fine-tune the Scene Flow pretrained model on the combined KITTI 2012 and KITTI 2015 datasets for 50,000 steps. 

\subsection{Comparisons with State-of-the-art}
We contrast the SOTA methods on Scene Flow \cite{sceneflow}, KITTI \cite{kitti2012,kitti2015}, and Middlebury \cite{middlebury} datasets. HART performs admirably on all of the datasets mentioned above.

\begin{table}[]
	\centering	
	\caption{Quantitative evaluation on Scene Flow test set,  the 1-pixel error rate is employed (lower is better). The \textbf{best} results are indicated in bold. $\star$ refers to Stereo Transformer methods.}
	\label{SF}
	\setlength{\tabcolsep}{1.0mm}{
		\begin{tabular}{l|c|c}
			\toprule[1.5pt]
			Method & Publish & EPE (px)$\downarrow$ \\
			\midrule
			GC-Net \cite{gcnet}& ICCV2017 & 1.84 \\
			GANet \cite{ganet}& CVPR2019 & 0.78  \\
			GwcNet \cite{gwcnet} & CVPR2019 & 0.76     \\
			AANet \cite{aanet}& CVPR2020 & 0.87  \\
			RAFT-Stereo \cite{raftstereo} & 3DV2021 & 0.61  \\
			ACVNet+ \cite{acvnet}& CVPR2022 &0.48	\\
			IGEV-Stereo \cite{igev} &  CVPR2023   & 0.47              \\
			Selective-IGEV \cite{selective} & CVPR2024 & \snd 0.44 \\ 
			\midrule
			STTR \cite{sttr} $\star$ &ICCV2021& \trd 0.45  \\ 
			DLNR \cite{dlnr} $\star$ &  CVPR2023    & 0.53              \\
			GOAT \cite{goat} $\star$  & WACV2024 & 0.47 \\ 
			RetinaStereo \cite{retina} $\star$  & ICASSP2025 & 0.49 \\
			\midrule
			HART (Ours) $\star$   & Ours & \fst \textbf{0.42}   \\     \bottomrule[1.5pt]
	\end{tabular}}	
\end{table}

\textbf{Scene Flow.} As shown in Tab. \ref{SF}, HART achieves SOTA performance on the Scene Flow dataset. It achieves a 6.67\% error reduction on the EPE metric compared to STTR \cite{sttr}.

\textbf{Reflective Area.} HART can integrate features to make more accurate matches in ill-posed regions, such as reflective area. HART ranked \textbf{1st} on the KITTI 2012 reflective benchmark at the time of submission, as shown in Tab. \ref{reflective2012}.

\begin{table*}[h]
	\centering
	\caption{Results on the KITTI 2012 Reflective leaderboard.  ``Noc" means percentage of erroneous pixels in non-occluded areas, ``All" means percentage of erroneous pixels in total.} 
\label{reflective2012}
\begin{tabular}{l|c|cccc|cccc|ccc}
	\toprule[1.5pt]
	\multirow{2}{*}{Method} & \multirow{2}{*}{Publish} &  & \multicolumn{2}{c}{D1\textgreater2px (\%)} &  &  & \multicolumn{2}{c}{D1\textgreater3px (\%)} &  &  & \multicolumn{2}{c}{D1\textgreater4px (\%)} \\ \cline{4-5} \cline{8-9} \cline{12-13} 
	&  &  & Noc & All &  &  & Noc & All &  &  & Noc & All \\ \midrule
	CREStereo \cite{cre} & CVPR2022 &  & 9.71 & 11.26 &  &  & 6.27 & 7.27 &  &  & 4.93 & 5.55 \\
	IGEV-Stereo \cite{igev}& CVPR2023 &  & 7.29 & 8.48 &  &  & 4.11 & 4.76 &  &  & 2.92 & 3.35 \\
	P3SNet+ \cite{p3snet}& TITS2023 &  & 23.60 & 26.32 &  &  & 15.85 & 18.50 &  &  & 11.66 & 14.13 \\
	UCFNet \cite{ucfnet}& TPAMI2023 &  & 9.78 & 11.67 &  &  & 5.83 & 7.12 &  &  & 4.15 & 4.99 \\
	GANet+ADL \cite{adl}& CVPR2024 &  & 8.57 & 10.42 &  &  & 4.84 & 6.10 &  &  & 3.43 & 4.39 \\
	Selective-IGEV \cite{selective} & CVPR2024 &  & \trd 6.73 & \snd 7.84 &  &  & \trd 3.79 & \snd 4.38 &  &  & 2.66 & \trd 3.05 \\
	LoS \cite{los} & CVPR2024 &  & \snd6.31 & \snd 7.84 &  &  & \snd3.47 & \trd 4.45 &  &  & \snd  2.41 & \snd  3.01 \\
	MoCha-Stereo \cite{mocha}& CVPR2024 &  & 6.97 & \trd 8.10 &  &  & 3.83 & 4.50 &  &  & \trd 2.62 & 3.80 \\
	ADStereo \cite{adstereo} & TIP2025 &  & 8.38 & 9.67 &  &  & 5.10 & 5.98 &  &  & 3.89 & 4.55 \\	
	\midrule
	HART & Ours &  & \fst \textbf{6.18} & \fst \textbf{7.38} &  &  & \fst \textbf{3.14} & \fst \textbf{3.92} &  &  & \fst \textbf{1.99} & \fst \textbf{2.49} \\
	\bottomrule[1.5pt]
\end{tabular}
\end{table*}

\begin{figure*}[]
	\centering
	\includegraphics[width=\linewidth]{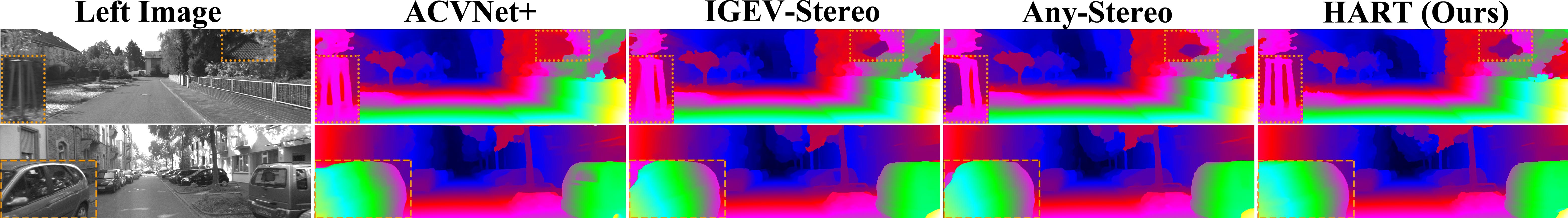}
	\caption{Visualisation on the KITTI dataset \cite{kitti2015}. We conducted comparisons with existing SOTA methods \cite{acvnet,igev,anystereo}. In the first image, HART accurately matches the details of the railing and correctly identifies the distant area between the tree leaves. In the second image, reflections on the car caused other methods to mis-match the outline of the car, while only HART achieved the correct match.
	} \label{KITTI2015fig}
\end{figure*}

\textbf{KITTI.} 
To evaluate the performance of HART in real-world scenarios, we conducted experiments on KITTI benchmarks. Stereo transformers exhibit a significant disadvantage in terms of time. This is primarily due to the quadratic time complexity of their self-attention. While DLNR achieves superior inference speed compared to other stereo transformers, this efficiency comes at the cost of compromised spatial dimension interaction. Consequently, the model exhibits constrained stereo matching accuracy in dynamic driving environments. HART overcomes this limitation and achieves improved performance under constrained inference time. Experimental results in Tab. \ref{2015} and Fig. \ref{KITTI2015fig} confirm that, HART is still able to emphasise the matching of spatial details within a given time.

\begin{table*}[]
	\centering
	\caption{Results on the KITTI 2015 and 2012. Error threshold is 3 px for 2015, and 4 px for 2012. Front-ground error is indicated by "F.G.", and Background error by "B.G.". $\star$ refers to Stereo Transformer methods, which are summarized in the lower half of the table. The upper half of the table describes Stereo CNN methods. ``Time" denotes the inference time on single NVIDIA Tesla A100.}
	\label{2015}	
	\setlength{\tabcolsep}{1.0mm}{
		\begin{tabular}{l|c|cccc|lccc|lccc|c}
			\toprule[1.5pt]			
			\multirow{2}{*}{Method} & \multirow{2}{*}{Publish} &  & \multicolumn{2}{c}{2015 All} &  &  & \multicolumn{2}{c}{2015 Noc} &  &  & \multicolumn{2}{c}{2012} &  & \multirow{2}{*}{Time (s)} \\ \cline{4-5} \cline{8-9} \cline{12-13}
			&  &  & F.G. & B.G. &  &  & F.G. & B.G. &  &  & Noc & All &  &  \\
			\midrule
			GwcNet \cite{gwcnet} & CVPR2019 && 3.93  & 1.74  &&& 3.49  & 1.61  &&& 0.99 & 1.27 && 0.32 \\
			ACVNet+ \cite{acvnet} &CVPR2022
			&& 3.07 &\fst  \textbf{1.37}  &&& 2.84 & \fst \textbf{1.26}  &&& \trd 0.86 & \trd 1.12 && 0.20 \\
			IGEV-Stereo \cite{igev} &CVPR2023
			&& 2.67 & \snd  1.38  &&& \trd 2.62 & \snd  1.27  &&& 0.87 & \trd 1.12 && 0.32 \\
			Any-Stereo \cite{anystereo} & AAAI2024 && 3.04 & 1.44  &&& 2.88 & 1.30  &&&  1.11 & 1.41 &&0.34 \\
			DKT-IGEV \cite{dkt} & CVPR2024 && 3.05 & 1.46  &&& 3.01 & 1.36  &&& - & - && 0.18 \\
			LoS \cite{los} & CVPR2024 && 2.81 & 1.42  &&& 2.66 & 1.29  &&& \snd  0.85 & \snd  1.06 && 0.19 \\
			\midrule		
			STTR \cite{sttr} $\star$ & CVPR2021 && 3.61 & 1.70  &&& - & -  &&& - & - && 0.78 \\	
			CroCo-Stereo \cite{croco} $\star$ & ICCV2023
			&& \trd 2.65  & \snd  1.38   &&& \snd  2.56 & 1.30  &&& - & - && 0.93 \\ 	
			DLNR \cite{dlnr} $\star$ &CVPR2023
			&& \snd  2.59  & 1.60  &&& - & -  &&& - & - && 0.33 \\ 
			FET \cite{fet}$\star$ & TCI2024 && -  & -   &&& 3.63 & 1.78  &&& - & - && - \\
			\midrule
			HART$\star$ & Ours && \fst \textbf{2.49} & \trd 1.39  &&& \fst \textbf{2.50} & \trd 1.29  &&& \fst \textbf{0.84} & \fst \textbf{1.05} &&0.25
			\\ \bottomrule[1.5pt]
	\end{tabular}}
\end{table*}

\begin{figure*}[]
	\centering
	\includegraphics[width=\linewidth]{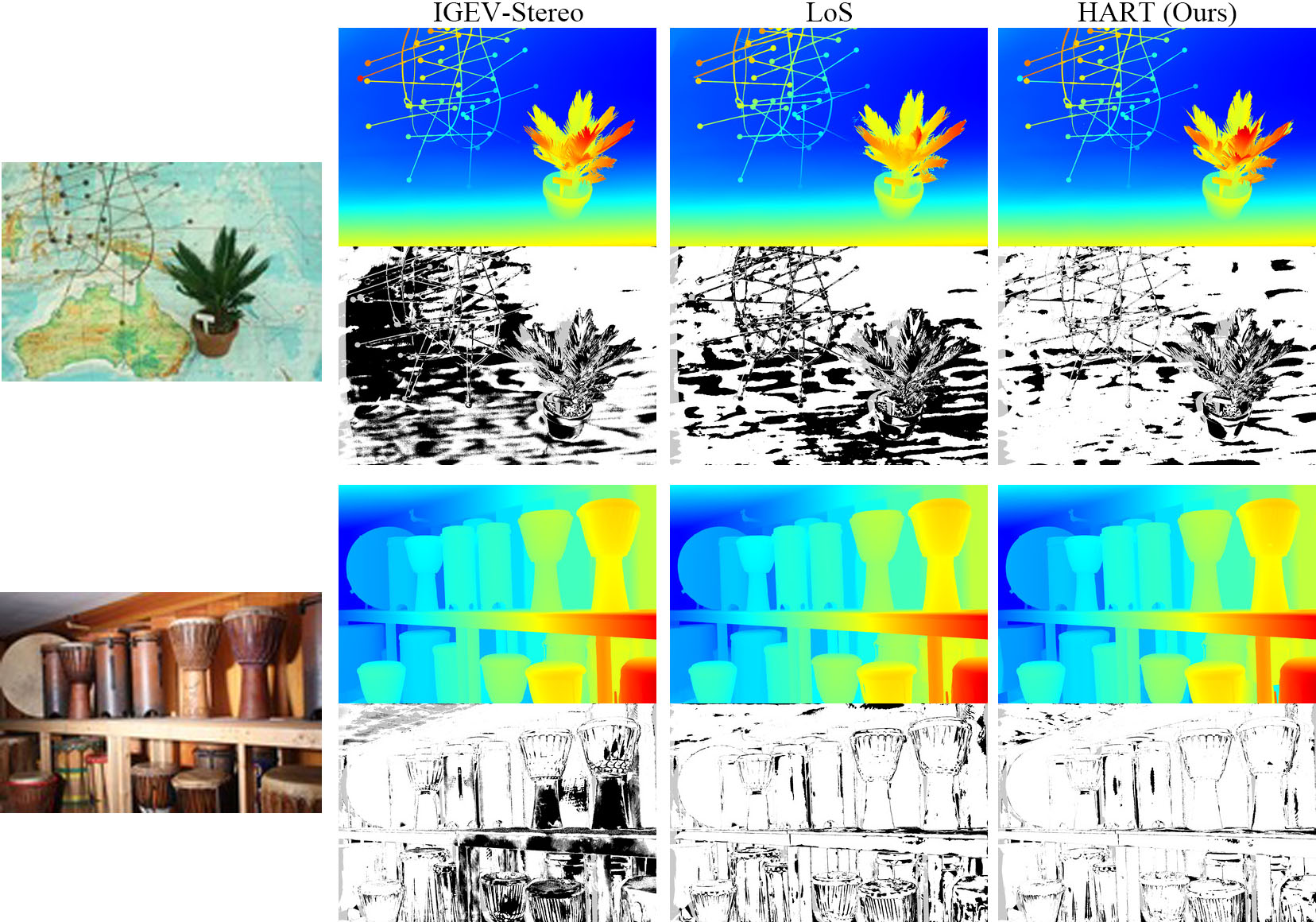}
	\caption{Visualisation on the Middlebury dataset \cite{middlebury}. Disparity and error maps at 0.5 px threshold are included. In the error maps, black regions indicate mismatched areas. We conducted comparisons with existing SOTA methods \cite{igev,los}. It can be observed that HART produces fewer mismatched regions compared to other methods.
	} \label{midd}
\end{figure*}

\textbf{Middlebury.} 
HART is also able to achieve excellent performance in the Middlebury scenario. HART outperforms the SOTA method LoS \cite{los} by \textbf{15.5\%}, \textbf{15.1\%}, \textbf{8.62\%} and \textbf{11.3\%} for the four metrics shown in Tab. \ref{mid_bench} and Fig. \ref{midd}.

\begin{table*}[]
	\centering
	\caption{Results on Middlebury. The error threshold are 0.5 px and 1 px here.}
	\label{mid_bench}
	\setlength{\tabcolsep}{1.0mm}{
		\begin{tabular}{l|c|lcclcc}
			\toprule[1.5pt]
			\multirow{2}{*}{Method}  & \multirow{2}{*}{Publish} &  & \multicolumn{2}{c}{D1$\textgreater$0.5px} &  & \multicolumn{2}{c}{D1$\textgreater$1.0px} \\ \cline{4-5} \cline{7-8} 
			&  &  & Noc & All &  & Noc & All \\ \midrule
			RAFT-Stereo \cite{raftstereo} & 3DV2021 &  & \snd 27.7 & \snd 33.0 &  & \trd 9.37 & 15.1 \\
			IGEV-Stereo \cite{igev} & CVPR2023 &  & 32.4 &  36.6 &  & 9.41 & \snd 13.8 \\
			Any-Stereo \cite{anystereo} & AAAI2024 &  & 32.7 & 37.5 &  & 11.5 & 16.8 \\
			GANet+ADL \cite{adl} & CVPR2024 &  & 52.3 & 55.9 &  & 30.3 & 35.2 \\
			LoS \cite{los} & CVPR2024 &  & \trd 30.3 & \trd 35.1 &  & \snd  9.05 &  \trd 14.2 \\	
			ADStereo \cite{adstereo} & TIP2025 & & 61.1 & 63.8 &  & 35.7 & 39.9 \\
			\midrule
			CSTR \cite{cstr} $\star$ & ECCV2022 &  & 48.5 & 55.2 &  & 21.6 & 32.0 \\
			CroCo-Stereo \cite{croco} $\star$ & ICCV2023 &  & 40.6 & 44.4 &  & 16.9 & 21.6 \\
			GOAT \cite{goat} $\star$ & WACV2024 &  & 49.0 & 52.9 &  & 21.4 & 27.0 \\
			\midrule
			HART $\star$ & Ours &  & \fst \textbf{25.6} & \fst \textbf{29.8} &  & \fst \textbf{8.27} & \fst \textbf{12.6}\\
			\bottomrule[1.5pt]
	\end{tabular}}
\end{table*}

\begin{figure*}[]	
	\centering
	\includegraphics[width=\linewidth]{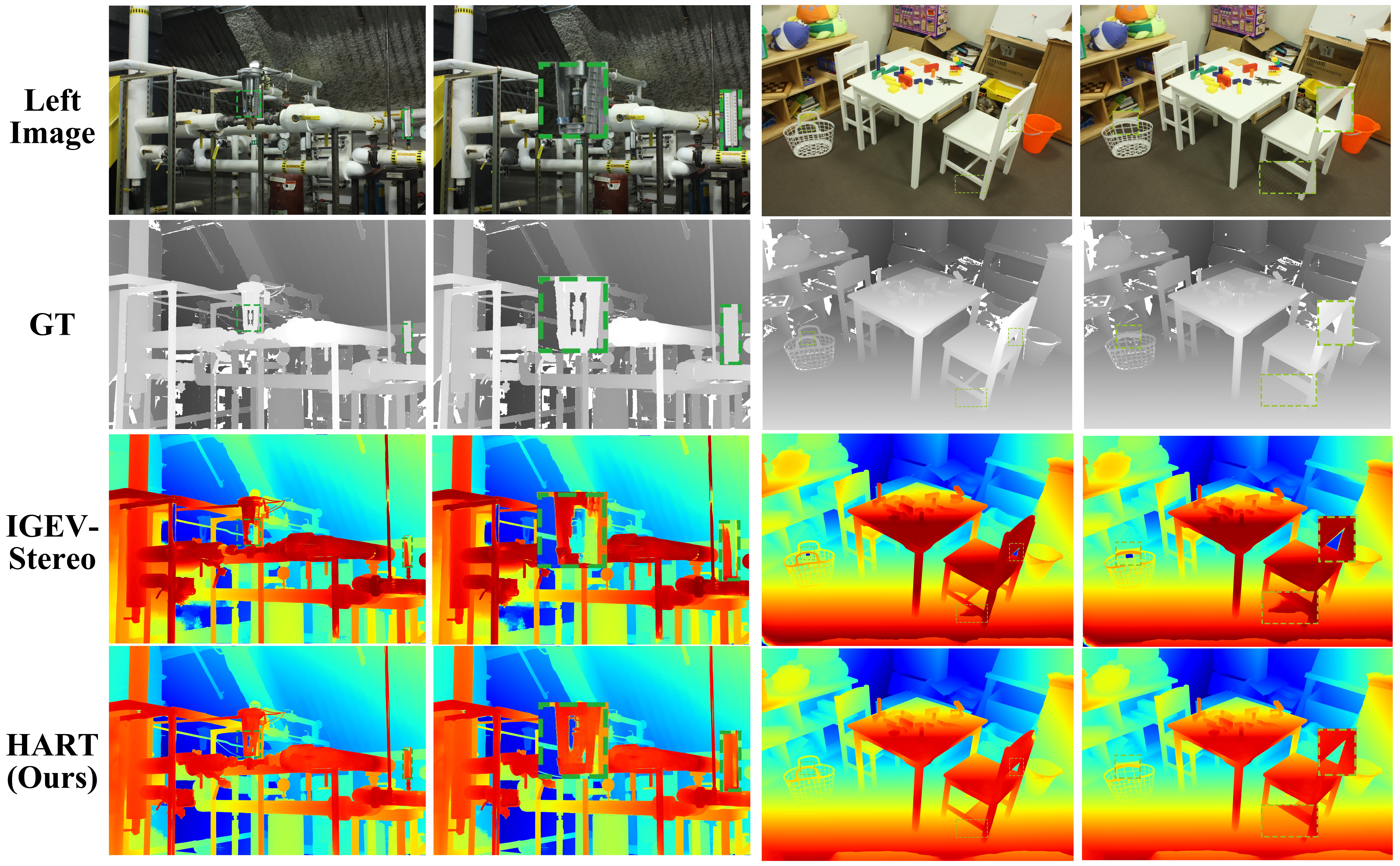}
	\caption{Zero-shot evaluation without fine-tuning on Middlebury. All results visualised here are trained on Scene Flow. The odd-numbered columns show the original images, the even-numbered columns present zoomed-in details for better visualization. In the first scenario, IGEV-Stereo misclassifies industrial equipment components as background and fails to recognize the thermometer as a co-planar object. In the second scenario, IGEV-Stereo is confused by thin objects, leading to incorrect disparity estimation in regions such as gaps within baskets and chairs.}
	\label{zero}
\end{figure*}

\begin{table*}[]
	\centering
	\caption{Zero-shot evaluation on Middlebury and KITTI. Every model undergoes scene flow training without fine-tuning. The 2px error rate is employed for Middlebury, and 3px error rate for KITTI. }
	\label{ZS}
	\begin{tabular}{l|c|cccccccc}
		\toprule[1.5pt]
		& && \multicolumn{3}{c}{Middlebury$\downarrow$} && \multicolumn{2}{c}{KITTI$\downarrow$} \\
		\cline{4-6} \cline{8-9}
		\multicolumn{1}{l|}{\multirow{-2}{*}{Method}}&\multicolumn{1}{c|}{\multirow{-2}{*}{Publish}}&& F & H & Q && EPE & D1 \\ 
		\midrule
		CFNet \cite{cfnet} &CVPR2021 && 28.2 & 15.3 & 9.8&& 1.7 & - \\
		RAFT-Stereo \cite{raftstereo} &3DV2021 && 23.4 & 12.6 & 7.3 && \trd 1.3 & 6.4 \\
		IGEV-Stereo \cite{igev} &CVPR2023 && 15.2 & \trd 7.1 & \trd 6.2 && \snd  1.2 & \trd 6.0 \\
		MoCha-Stereo \cite{mocha} &CVPR2024 && \snd  12.4 & \snd  6.2 & \snd  4.9 && \snd  1.2 & \snd  5.8 \\ \midrule
		STTR \cite{sttr} $\star$ &ICCV2021 && - &15.5&9.7 && 1.5 & 6.4 \\			
		CSTR \cite{cstr} $\star$ &ECCV2022 && - & - & - && 1.4 & \snd  5.8 \\
		ELFNet \cite{elfnet} $\star$ &ICCV2023 && - & 10.0 & 7.9 && 1.6 & \snd  5.8 \\
		DLNR \cite{dlnr} $\star$ &CVPR2023 && \trd 14.5 & 9.5 & 7.6 && 2.8 & 16.1 \\
		FET \cite{fet} $\star$ & TCI2024 && - & - & - && 1.4 & 6.7 \\
		\midrule
		HART $\star$ & Ours && \fst \textbf{12.2} & \fst \textbf{4.9} & \fst \textbf{4.8} && \fst \textbf{1.1} & \fst \textbf{5.1} \\
		\bottomrule[1.5pt]
	\end{tabular}
\end{table*}

\subsection{Zero-shot Generalization}
Collecting stereo matching datasets poses significant challenges, and stereo matching in real-world scenes often requires training on a small dataset to obtain matching results in untrained real scenes. This places certain demands on the generalization performance of stereo matching models. Zero-shot generalization is a common experiment to test the generalization performance of stereo matching. We follow the paradigm of training on Scene Flow and testing the model on real scenes from Middlebury, KITTI, and Driving Stereo \cite{drivingstereo} datasets to validate our generalization ability. The experimental results are presented in Tab. \ref{ZS}, Tab. \ref{ds_zs} and Fig. \ref{zero}. The quantitative comparison results demonstrate the excellent generalization performance of HART. 

\begin{table*}[]
	\centering
	\caption{Zero-shot evaluation on the Driving Stereo dataset. The metric used is the End Point Error (EPE) on the full-resolution (F) and half-resolution (H) images.}
	\label{ds_zs}
	\begin{tabular}{l|c|cccccccccccc}
		\toprule[1.5pt]
		\multirow{2}{*}{Method} & \multirow{2}{*}{Publish} & & \multicolumn{2}{c}{Sunny} &  & \multicolumn{2}{c}{Rainy} &  & \multicolumn{2}{c}{Cloudy} &  & \multicolumn{2}{c}{Foggy} \\ \cline{4-5} \cline{7-8} \cline{10-11} \cline{13-14} 
		& & & F & H &  & F & H &  & F & H &  & F & H \\ \midrule
		DLNR \cite{dlnr} &CVPR2023 & & 3.16 & 1.93 &  & \trd 4.94 & 3.85 &  & 2.45 & \trd 1.85 &  & 3.47 & 2.63 \\		
		MoCha-Stereo \cite{mocha}& CVPR2024& & \snd  1.92 & \trd 1.19 &  & \snd  4.21 & \trd 2.37 &  & \snd  1.93 & \fst \textbf{0.99} &  & \snd  2.00 & \snd  0.98 \\
		Selective-IGEV \cite{selective}& CVPR2024& & \trd 2.18 & \snd  1.18 &  & 5.46 & \snd  2.17 &  & \trd 2.15 & \snd  1.12 &  & \trd 2.32 & \trd 1.10 \\  \midrule
		HART & Ours & & \fst \textbf{1.74} & \fst \textbf{1.02} &  & \fst \textbf{2.97} & \fst \textbf{1.40} &  & \fst \textbf{1.70} & \fst \textbf{0.99} &  & \fst \textbf{1.74} & \fst \textbf{0.94}\\ 
		\bottomrule[1.5pt]
	\end{tabular}
\end{table*}

\subsection{Ablations}
To validate the effectiveness of our design, we conducte ablation studies by individually removing each module. All experiments are conducted on the Scene Flow \cite{sceneflow} dataset, and all hyperparameters are kept consistent with \cite{igev}. The results of the ablation experiments are shown in Tab. \ref{abl}. In addition to standard metrics such as End Point Error (EPE) and D1, we also evaluate the $\frac{Rank(A)}{m}$ metric. This new metric computes the average percentage of the ratio between the rank of the attention matrix and its dimension $m$ (i.e., the number of rows in the attention matrix) across inference steps. 

\textbf{Baseline No.1: Recurrent Stereo CNN.} 
Our No.1 baseline is IGEV-Stereo \cite{igev}. Experiment No.1 in Tab. \ref{abl} demonstrates the performance of this method. Our transformer decoder borrows its code related to iterative updating and correlation calculation.

\begin{table*}[]
	\centering
	\caption{Ablation study. Error threshold is 3 px. ``\textbf{ID}" means \textbf{I}GEV \textbf{D}ecoder. The following metrics represent the results obtained from training and testing on the Scene Flow dataset. $\frac{Rank(A)}{m}$ computes the average percentage of the ratio between the rank of the attention matrix and its dimension $m$ (the number of rows in the attention matrix) across inference steps. ``Time" denotes the inference time on single NVIDIA Tesla A100.}
	\label{abl}
	\setlength{\tabcolsep}{0.3mm}{
		\begin{tabular}{c|cccccccc|c|cc|c}
			\toprule[1.5pt]
			\multirow{2}{*}{No.} && STTR & \multicolumn{2}{c}{HPSA Encoder} &&& \multirow{2}{*}{ID} && $\frac{R(A)}{m}$ & EPE & D1 & Time \\
			\cline{4-7}
			&& Encoder & Attention  & MKOI &&&   & & (\%) & (px) & (\%) & (s) \\	\midrule
			1&&  \multicolumn{8}{c}{\cellcolor{gray} CNN architecture}  & \multicolumn{1}{|c}{0.47} & 2.47 & 0.37 \\	
			2&& \checkmark & Vanilla SA & - &&& \checkmark && 78.1 & 0.44 & 2.38 & 0.71 \\	\midrule
			3&& - & Hadamard product w/ SoftMax & - &&& \checkmark &&81.3& 0.49 & 2.46 & \textbf{0.28} \\
			4&& - & Hadamard product w/ DAK & - &&& \checkmark &&93.0& 0.46 & 2.43 & \textbf{0.28} \\
			5&& - & Hadamard product w/ SoftMax & \checkmark &&& \checkmark &&81.8& 0.43 & 2.39 & 0.36 \\
			\midrule
			6&& - & Hadamard product w/ DAK  & \checkmark &&& \checkmark &&94.5& \textbf{0.42} & \textbf{2.32} & 0.36 \\
			\bottomrule[1.5pt]
	\end{tabular}}
\end{table*}

\textbf{Baseline No.2: Recurrent Stereo Transformer.} 
Building upon STTR \cite{sttr}, we designed the STTR encoder with the IGEV Decoder, as introduced in Section \ref{igevdecoder}. 
The improved version of STTR (No. 2 Experiment) has achieved higher accuracy and faster inference time compared to the original STTR \cite{sttr}. This improvement is attributed not only to the enhanced decoder, i.e., IGEV Decoder, but also to the adoption of techniques inspired by Restormer \cite{restormer}. Specifically, the feed-forward propagation network of the encoder has been upgraded to the SGFF structure introduced in Section \ref{sgff}. 
Through this combination, we obtained a Recurrent Stereo Transformer. 
The effectiveness of this Recurrent Stereo Transformer, which is based on the vanilla self-attention (SA), is showcased in Experiment No.2. 

Despite achieving results surpassing the CNN paradigm using the vanilla attention mechanism, the square complexity of matrix multiplication significantly slowed down the inference time, rendering it unsuitable for efficient matching requirements in practical scenarios. To address this issue, we propose an improved version of transformer encoder called HPSA, and utilize a series of novel designs to ensure its performance. 

\textbf{Hadamard product between Query and Key.} Hadamard product is the foundation of our HPSA. By utilizing the Hadamard product for attention computation, HART achieves linear attention computation. The effectiveness of this caculation is demonstrated in Experiment No.3. Although combining HP with the vanilla SoftMax activation function significantly reduces inference time, this approach encounters two issues. 1) The limitation of the SoftMax activation function prevents HART from overcoming the low-rank bottleneck \cite{lowrank} problem. 2) The Hadamard product lacks spatial and channel interactions, making it challenging to achieve satisfactory performance solely through the use of the Hadamard product.

\begin{figure}[h]		
	\centering
	\includegraphics[width=0.5\linewidth]{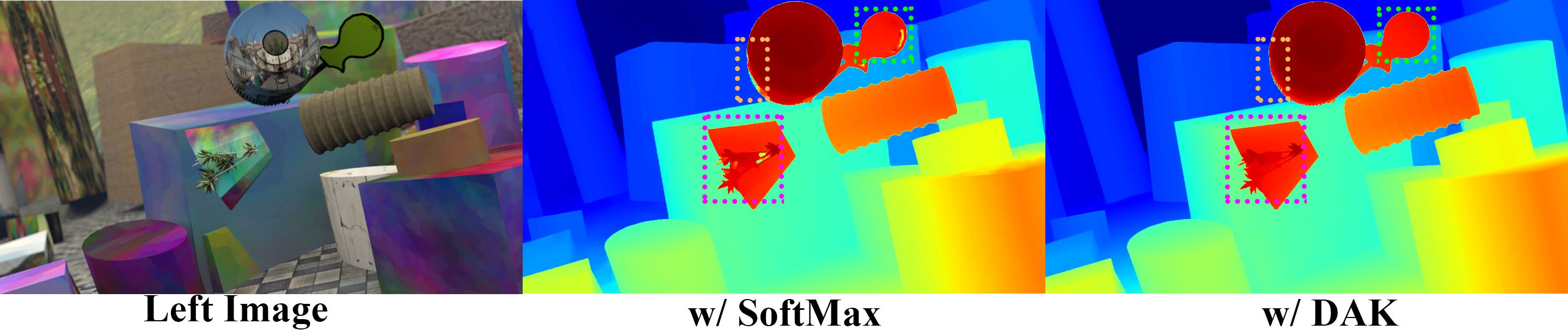}
	\caption{Visualization of the effects using DAK and SoftMax as kernel functions is presented. The figure shows a visualization of the model trained on the Scene Flow Train dataset, evaluated on a sample scene from the Scene Flow Test dataset. As highlighted in the zoomed-in region, using DAK as the kernel function leads to improved matching performance in ill-posed regions such as reflective surfaces, edge textures, and thin objects. 
	}
	\label{abl-dak}
\end{figure}

\textbf{Dense Attention Kernel (DAK).}
The SoftMax operation converts attention weights into a probability distribution through exponential normalization. However, this process excessively suppresses features at certain positions, driving some weights in the attention matrix close to zero and consequently leading to loss of features. As shown in Fig. \ref{abl-dak} Such sparsity may impair the model's ability to capture contextual information in ill-posed region, resulting in a low-rank bottleneck. DAK is proposed to address the low-rank bottleneck issue caused by SoftMax. The DAK ensures that the attention matrix remains positive, enabling effective focus on ill-posed regions. As shown in Tab. \ref{abl}, from the perspective of matrix rank, DAK contributes to achieving higher ranks, which to some extent enhances the model's representational capacity. Comparative experiments No.3 and No.4 demonstrate the effectiveness of DAK.

\begin{table*}[h]
	\centering
	\caption{Ablation studies on the MKOI module. Experiment No. 1 corresponds to the default configuration of MKOI. The table under "Value Matrix" summarizes the convolution operations along with their corresponding input and output channel dimensions.}
	\label{abl_mkoi}
		\begin{tabular}{l|c|ccccc|cc|c}
			\toprule[1.5pt]
			\multirow{2}{*}{No.} & \multirow{2}{*}{Attention Matrix} &  & \multicolumn{3}{c}{Value Matrix}      &  & \multirow{2}{*}{EPE} & \multirow{2}{*}{D1} & \multirow{2}{*}{Para. (M)} \\ \cline{4-6}
			&                                   &  & 3$\times$3     & 5$\times$5     & 7$\times$7    &  &                      &                     &                            \\ \midrule
			1                    & \multirow{3}{*}{1$\times$1 Conv, $\frac{7}{4}$C}     &  & C        & C/2     & C/4     &  & \textbf{0.42}                 & 2.32                & 107.27                         \\
			2                    &                                   &  & C        & C       & C        &  &       0.43         & \textbf{2.31}                   & 163.13                         \\
			3                    &                                   &  & C/4        & C/2       & C        &  &       0.45         & 2.48                  & 151.01                         \\ \midrule
			4                    & 7$\times$7 Conv, $\frac{7}{4}$C                      &  & C        & C/2         & C/4    &  & 0.43                    &       2.40            &  170.68                         \\ 
			5                    & 1$\times$1 Conv, $\frac{7}{4}$C Only                 &  & \multicolumn{3}{c}{3$\times$3 Conv, $\frac{7}{4}$C Only} &  &        0.46             &    2.44     & 96.12             \\
			\bottomrule[1.5pt]
		\end{tabular}
\end{table*}

\textbf{Multi Kernel \& Order Interaction (MKOI).}
The HPSA without MKOI can be represented by Equ. \ref{wo}. 
\begin{eqnarray}
HPSA(Q,K,V)_{w/o~MKOI}=DAK(||Q||_2\odot ||K||_2)\odot V
\label{wo}
\end{eqnarray}
Since the interaction between matrices is performed entirely through Hadamard product, Equ. \ref{wo} fails to facilitate spatial and channel dimensional interactions. We further developed MKOI to overcome this shortcoming. Experiments No.5 and No.6 demonstrate the performance of MKOI.

We argue that semantic information has limited correlation with fine-grained details, and thus full feature channels are not necessary to achieve comparable performance. Consequently, we apply large-kernel convolutions only on a subset of the channels. Experiments No. 1 and No.2 employ 1$\times$1 convolutions to expand the feature channels of the attention matrix from C to $\frac{7}{4}$C. As shown in Tab. \ref{abl_mkoi}, when the channel dimension for large-kernel convolution is reduced to C/4, the number of parameters decreases while maintaining equivalent accuracy.

Local detail features require sliding convolution with small kernels across all channels.
Experiment No. 3 adopts a design logic opposite to that of MKOI: large-kernel convolutions are applied to all feature channels, while small-kernel convolutions are used on a subset of channels. Experiment No. 4 utilizes a 7$\times$7 convolution to upscale the feature channels of the attention matrix from C to $\frac{7}{4}$C. These results demonstrate that using only a subset of channels or applying large-kernel convolutions on the attention matrix fails to achieve satisfactory performance. 

Furthermore, experiment No. 5 employs only standard 1$\times$1 convolutions: first, to expand the feature channels of both the attention and value matrices from C to $\frac{7}{4}$C; then, their Hadamard product is passed through another 1$\times$1 convolution to reduce the channel dimension back to C, producing the final output of the attention process. The comparison with standard convolution further demonstrates the effectiveness of MKOI, and confirms that MKOI’s paradigm—splitting channels first and then performing interaction—is effective.

\begin{table}[h]
	\centering
	\caption{Ablation study for iterations. The metric here is end point error} 
	\label{iter}	
	\setlength{\tabcolsep}{0.5mm}{
		\begin{tabular}{ccccccc}
			\toprule[1.5pt]
			& \multicolumn{6}{c}{Number of Iterations} \\ \cline{2-7} 
			& 1    & 2    & 3    & 4    & 8    & 32   \\	\midrule
			RAFT-Stereo \cite{raftstereo} &   2.08   &  1.13    &  0.87    &   0.75   &   0.58    &   0.53   \\	
			IGEV-Stereo \cite{igev} &  \trd 0.66   & \trd 0.62    & \trd 0.58    &  \trd 0.55   &  \trd 0.50    &  \trd 0.47   \\	
			Selective-IGEV \cite{selective} &  \snd 0.65   & \snd 0.60   &  \snd 0.56    &  \snd 0.53   &  \snd 0.48    &  \snd 0.44   \\ \midrule
			DLNR \cite{dlnr} $\star$&  1.56 & 0.96 & 0.80 & 0.71 & 0.58 & 0.53 \\
			RetinaStereo \cite{retina} $\star$&  1.00 & 0.80 & 0.69 & 0.62 & 0.52 & 0.49 \\
			\midrule
			HART (Ours) $\star$ &  \fst \textbf{0.60}   &  \fst \textbf{0.55}   &  \fst \textbf{0.51}   &  \fst \textbf{0.48}   &  \fst \textbf{0.45}    & \fst \textbf{0.42}  \\
			\bottomrule[1.5pt]
	\end{tabular}}
\end{table}

\textbf{Number of Iterations}
We adopted standard parameter settings and conducted our tests with 32 iterations. The architecture of HART enhances the effectiveness of the iterative strategy. HART achieves superior performance compared to other recurrent-based stereo CNNs \cite{raftstereo,igev,selective} and stereo transformers \cite{dlnr,retina} across various iteration counts. To substantiate this claim, we compared HART against the SOTA Recurrent Stereo Transformer \cite{dlnr} and Recurrent Stereo CNN \cite{igev}. As shown in Tab. \ref{iter}, experiments demonstrate that HART consistently outperforms other recurrent-based methods across various iteration counts.

\begin{figure}[]		
\centering
\includegraphics[width=\linewidth]{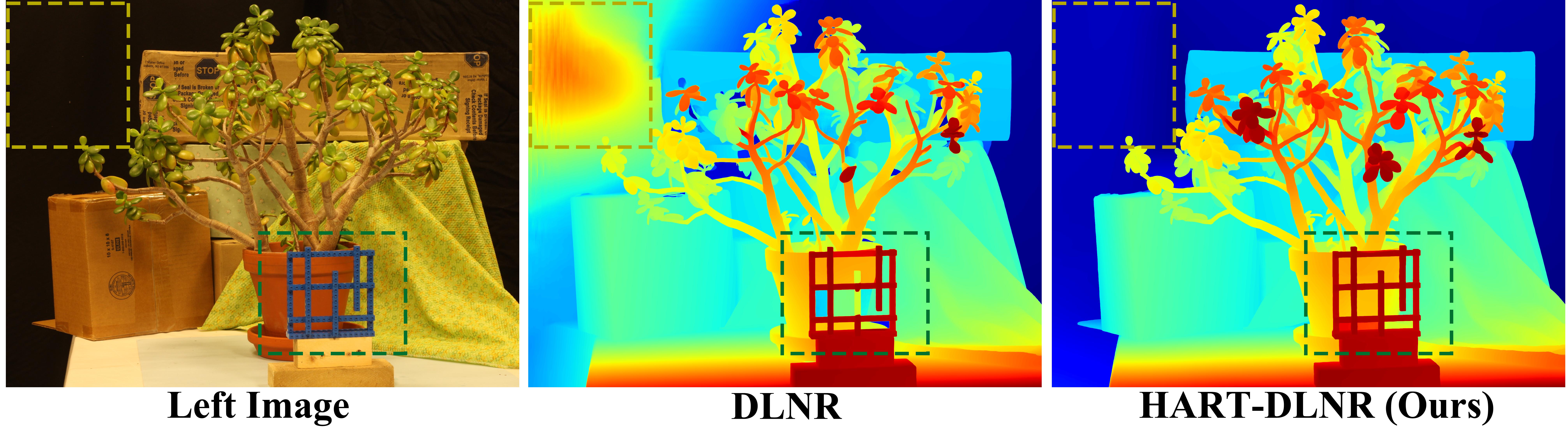}
\caption{Visualisation of the zero-shot performance comparison between \cite{dlnr} and our HART-DLNR. The results show that our HPSA can achieve better performance. 
}
\label{vsDLNR}
\end{figure}

\begin{table}[]
	\centering
	\caption{Zero-shot performance of HART-DLNR. The baseline here is DLNR. Our proposed HPSA Is used to replace self-attention mechanism (CATE) of DLNR. All models are trained on Scene Flow, and tested on Middlebury and KITTI. The threshold is 2px for Middlebury, and 3px for KITTI. ``Time (s)" is the inference time required for the model to run on the KITTI dataset using the NVIDIA A100. }
	\label{HART-DLNR}
	\setlength{\tabcolsep}{0.0 mm}{
		\begin{tabular}{c|c|lcccl|c}
			\toprule[1.5pt]
			\multirow{2}{*}{No.} & \multirow{2}{*}{Attention Mechanism} &&  \multirow{2}{*}{Middlebury$\downarrow$} & \multicolumn{2}{c}{KITTI$\downarrow$} && \multirow{2}{*}{Time (s)$\downarrow$} \\
			\cline{5-6}
			& & && EPE & D1 &&  \\ \midrule
			1&CATE \cite{dlnr}  && 14.5 & 2.8 & 16.1 && 0.33 \\ \midrule
			2&HP w/ SoftMax && 18.9 & 2.2 & 6.7 && 0.20 \\
			3&HP w/ DAK && 14.8 & 1.7 & 6.2 && \textbf{0.19} \\ 
			4&HPSA w/ SoftMax+MKOI&& 12.8 & 1.6 & 5.8 && 0.28 \\
			\midrule
			\multirow{2}{*}{5}&HPSA w/ DAK+MKOI&&  \multirow{2}{*}{\textbf{9.9}} & \multirow{2}{*}{\textbf{1.4}} & \multirow{2}{*}{\textbf{5.0}} && \multirow{2}{*}{0.27} \\
			&(full model of HART-DLNR) &&  &  &  &&  \\
			\bottomrule[1.5pt]
	\end{tabular}}
\end{table}

\subsection{Extension to existing Stereo Transformer}
To further validate that our attention paradigm is a better way for stereo transformers, we applied HPSA from our HART to replace the Channel-Attention Transformer Extractor (CATE) from DLNR. 
DLNR is a representative Recurrent Stereo Transformer. Although CATE makes less inference time than other stereo transformers \cite{elfnet,croco}, the trade-off is the loss of classical spatial interactions achieved by traditional ViTs.
What is more, it still suffers from common issues inherent to stereo transformers. Firstly, CATE still exhibits quadratic time complexity, leaving room for improvement in inference time. Additionally, the parameter count for each attention head in CATE is still $2nc/d$, which remains significantly smaller than $c^2$. This implies that DLNR also faces the issue of low-rank bottleneck \cite{lowrank}. Compared to CATE in DLNR, HPSA offers the following advantages: 1) Further reduction in computational complexity; 2) Alleviating the low-rank bottleneck by using DAK; 3) Spatial dimension interaction has been achieved; 4) Improve the way channel dimensions interact. These benefits address the limitations of CATE and allow HPSA to replace CATE and achieve better outcomes. Experimental results shown in Fig. \ref{vsDLNR} and Tab. \ref{HART-DLNR} demonstrate that HPSA designs improve the performance of DLNR and reduce the inference time.
\section{Conclusion}
HART introduces a novel design of the transformer architecture. The goal is to improve the stereo matching capability of transformers in ill-posed regions while overcoming the limitations imposed by quadratic computational complexity. DAK is established to solve the intrinsic constraints of low-rank bottleneck. MKOI is proposed to complement the spatial and channel interaction capabilities. 
The overall pipeline controls the time complexity within $O(n)$ using an attention pipeline based on Hadamard product. 
These designs allow HART to achieve SOTA performance. Experimental results demonstrate the effectiveness of our HART. HART ranks \textbf{1st} on the KITTI 2012 reflective benchmark and \textbf{TOP 10} on the Middlebury benchmark among all published methods. While HART still has its limitation. HART is not able to achieve real-time at the moment. In the future, we plan to implement a faster solution and validate the effectiveness of HART in more frameworks.




\section*{Declaration of competing interest}
\par{The authors declare that they have no known competing financial interests or personal relationships that could have appeared to influence the work reported in this paper.}

\section*{Funding}
\par{This work was supported in part by the Science and Technology Planning Project of Guizhou Province, Department of Science and Technology of Guizhou Province, China under Grant QianKeHe [2024] Key 001; and in part by the Science and Technology Planning Project of Guizhou Province, Department of Science and Technology of Guizhou Province, China under Grant [2023]159.}
\bibliographystyle{unsrt}
\bibliography{ref}

\end{document}